\documentclass{article}

\usepackage{arxiv}

\usepackage[utf8]{inputenc} % allow utf-8 input
\usepackage[T1]{fontenc}    % use 8-bit T1 fonts
\usepackage{hyperref}       % hyperlinks
\usepackage{url}            % simple URL typesetting
\usepackage{booktabs}       % professional-quality tables
\usepackage{amsfonts}       % blackboard math symbols
\usepackage{nicefrac}       % compact symbols for 1/2, etc.
\usepackage{microtype}      % microtypography
\usepackage{graphicx}
\usepackage{natbib}
\usepackage{doi}
\usepackage{amssymb}
\usepackage{amsmath}
\usepackage{amsthm}
\usepackage{booktabs}
\usepackage{algorithm}
\usepackage{algorithmic}
\usepackage[switch]{lineno}
\usepackage{latexsym}
\usepackage{cleveref}  
\usepackage{color}
\usepackage{comment}

\newtheorem{theorem}{Theorem}

\newtheorem{property}[theorem]{Property}

\newtheorem{definition}{Definition}

\title{Ranking counterfactual explanations }

\usepackage{authblk}

\newbox{\orcid}\sbox{\orcid}{\includegraphics[scale=0.06]{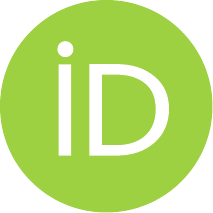}} 
\author[1]{%
	\href{https://orcid.org/0000-0000-0000-0000}{\usebox{\orcid}\hspace{1mm}Suryani Lim\thanks{\texttt{slim@federation.edu.au}}}%
}
\author[2]{%
	\href{https://orcid.org/0000-0000-0000-0000}{\usebox{\orcid}\hspace{1mm}Henri Prade\thanks{\texttt{henri.prade@irit.fr}}}%
}
\author[2]{%
	\href{https://orcid.org/0000-0000-0000-0000}{\usebox{\orcid}\hspace{1mm}Gilles Richard\thanks{\texttt{gilles.richard@irit.fr}}}%
}
\affil[1]{Federation University, Australia}
\affil[2]{IRIT-CNRS Toulouse, France}

\renewcommand{\shorttitle}{Ranking counterfactual explanations}

\begin{document}
\maketitle
\begin{abstract}
AI-driven outcomes can be challenging for end-users to understand.
Explanations can address two key questions: "Why this outcome?" (factual) and "Why not another?" (counterfactual).
While substantial efforts have been made to formalize factual explanations, a precise and comprehensive study  
of counterfactual explanations is still lacking.
This paper proposes a formal definition of counterfactual explanations, proving some properties they satisfy, and examining the relationship with factual explanations. 
Given that multiple counterfactual explanations generally exist for a specific case, we also introduce a rigorous method to rank these counterfactual explanations, going beyond a simple minimality condition, and to identify the optimal ones. Our experiments with 12 real-world datasets highlight that, in most cases, a single optimal counterfactual explanation emerges. 
We also demonstrate, via three metrics, that the selected optimal explanation exhibits higher representativeness and can explain a broader range of elements than a random minimal counterfactual. This result highlights the effectiveness of our approach in identifying more robust and comprehensive counterfactual explanations. 
\end{abstract}
\section{Introduction}\label{intro}
In the field of eXplainable AI (XAI), extensive research has focused on providing human-understandable explanations for decisions made by AI systems. Typically, this is achieved by analyzing global patterns within the data or examining local neighbors of a specific instance to offer factual explanations that address questions like, ``Why does this decision occur in this case?". From a logical perspective, such explanations are referred to as {\it abductive}. 
These explanation methods can be categorized as either model-agnostic or model-specific, depending on whether they can explain the decisions of any classifier or are tailored to a specific one. Additionally, they can also be classified as local or global, based on whether they explain the rationale behind a single decision or provide insights into the overall logic of a classifier.

More recently, counterfactual (or contrastive) explanations have gained attention for their ability to shed light on decisions by answering questions like  ``Why, in this case, was this decision not made instead?" (see \cite{HoyGenPsychoBulletin2017,WacMitRusHarvard2018,MillerAI2019}). In this context, the explanation consists of a contrastive example (i.e. an observable instance leading to a different decision) indicating what changes are needed to alter the outcome. Ultimately, only the pairs (feature, value) to be modified can be considered as counterfactual explanations.
Extracting counterfactual explanations relates to the broader challenge of understanding data, whether it comes from an AI-driven process or a 
%straightforward 
direct data collection procedure. To streamline the text,  we  may use ``counterfactual" in place of ``counterfactual explanation".

While a significant amount of works have been devoted to developing axiomatic definitions for factual explanations, no such effort has been made for counterfactual explanations. In this paper, we propose several definitions to capture the desirable properties of counterfactual explanations. We also investigate these properties and their relationship to factual explanations. Since multiple counterfactuals may exist for a single case, we also introduce a method to rank them, enabling the identification and presentation of a unique optimal counterfactual. 
Unlike current definition of the ``optimal counterfactual" which define optimal counterfactual explanation by minimizing only the Hamming distance~\cite{{KeaSmyICCBR2020,bouma2022}}, we incorporate the concept of ``counterfactual power", where the optimal counterfactual is the one with the highest generality power. 
Additionally, we introduce two metrics, typicality and universality to evaluate the quality of a counterfactual.
Our experiments on 12 real-world datasets demonstrate that in over 80\% of cases, a unique optimal counterfactual can be identified. Also, regarding the two metrics, the optimal counterfactual appears to be of better quality than a random one.

Our key contributions are:
%\begin{enumerate}
%\item 

1. We propose a formal definition of counterfactual explanation as opposed to factual explanations.
%\item 

2. We investigate the properties of counterfactuals deduced from the definition and we examine the link with factual explanations. 
%\item 

3. We propose a way to rank the counterfactuals, leading to the concept of optimal counterfactual explanations.
%\item 

4. We define three metrics, namely  typicality, capacity and universality, to provide a more robust assessment of the counterfactuals' quality.

5. We evaluate the effectiveness of our counterfactual definitions with 12 publicly available datasets. 

The paper is structured as follows: Section \ref{related} reviews related work. Section \ref{background} introduces the notations and concepts used throughout the paper, including the definitions of factual and counterfactual explanations. In Section \ref{properties}, we investigate the properties of counterfactual explanations and examine their link with factual explanations. Section \ref{ranking} presents a method for ranking counterfactuals for a given instance, leading to the definition of an optimal counterfactual. 
We also introduce three metrics to evaluate the quality of a counterfactual: typicality, capacity and universality.
Section \ref{expes} presents our experiments, describing the datasets used and analyzing the results. All code is publicly available at https://gitlab.com/pubcoding/arxiv2025.
Finally, Section \ref{conc} concludes the paper with future directions and closing remarks.
\section{Related Work}\label{related}
Among the various explanations for a given outcome, significant attention has been dedicated to factual explanations addressing the question, `Why this outcome?'. Early research focusing on the stability of neural networks under small perturbations to their inputs \cite{SzeZaretAlICLR2014}, introduced the concept of adversarial examples. An adversarial example arises from a minimal perturbation (measured with respect to an appropriate metric) of the input, resulting in a different output from the neural network. This phenomenon has serious implications, as it can enable malicious attacks that are imperceptible to humans but force the network to produce an altered result \cite{GooShlSzeICLR2015,ChaDeyChaMukarXiv2018}.
Ultimately, an adversarial example can be regarded as a specific type of counterexample. From the first-order logic perspective, there is a deep connection between explanations and adversarial examples: counterexamples can be generated from factual explanations and vice versa~\cite{JoaoNEURIPS2019,JoaoArxiv2020}.
The growing interest in counterfactual explanations is unsurprising, as they offer a compelling alternative or complement to traditional explanation systems. For example, motivated by the European GDPR \cite{GDPR2017}, \cite{WacMitRusHarvard2018} developed a comprehensive system that generates only counterfactual explanations. But other systems, such as LORE \cite{GuidottiLORE2018,GuiMonGiaPedRugTurIEEEIS2019}, provide both factual and counterfactual explanations.

In \cite{DhurandharetAlNIPS2018}, the authors introduced the `Contrastive Explanations Method' (CEM), tailored for neural networks, which identifies not only the relevant positive features (those supporting the outcome) but also the pertinent negative features—those whose absence would prevent the outcome from changing. This aligns with recommendations outlined in \cite{DoshiaretAlXiv2017}. Counterfactual explanations have even been applied to image classification. In~\cite{taylor2024}, the authors proposed a causal generative explainer to help end-users understand why a certain digit in the Morpho-MNIST causal dataset was classified as another digit. Finally, in \cite{LimPraRicECAI2024}, the authors propose a counterfactual explainer based on analogical proportions. 

Notably, while all these works generate multiple counterfactual explanations for a given outcome, they generally do not propose a systematic method for ranking these explanations beyond ensuring minimality. We address this gap in the present paper.
Our approach is agnostic to specific machine learning models and processes, making it broadly applicable across categorical datasets as a robust method for counterfactually explaining outcomes.

\section{Background}\label{background}
We provide the general notation and definitions for the following subsections.
\subsection{Basic definitions and notations}
We consider $n$ categorical features indexed from $1$ to $n$, where each feature takes values from a finite set $X_i$. A literal is defined as a pair $(j,v)$, where $j \in [1,n]$ and $v \in X_j$, representing the condition ``feature $j$ has value $v$".

Let $X=X_1 \times \ldots \times X_n$. The set of all possible literals constructed from $X$ is denoted by $Lit(X)$. 
A set of literals $\psi$ is called consistent if it includes at most one literal for each feature; otherwise, it is considered inconsistent (i.e.,  providing 2 different values for the same feature). Consequently, if $\psi$ is consistent, then $|\psi| \leq n$ (where $|.|$ denotes the cardinality of a set).
The set of all consistent sets of literals is denoted by $Cons(X)$, which is a subset of the power set of $Lit(X)$, i.e., 
$Cons(X) \subset 2^{Lit(X)}$.

An instance is defined as a consistent set of literals with cardinality $n$. The space of all such instances is denoted by $Inst(X)$, and $Inst(X) \subset Cons(X)$.
Frequently, the whole instance space is  only observable through a sample $S$, primarily due to the huge size of $Inst(X)$ in high dimensions. Alternatively, there may exist combinations of features that cannot represent a real world item, such as when employing one-hot encoding. Consequently, $Inst(X)$ is more of a theoretical concept than a practical one.

In this setting, a real world item is represented as an instance $a=\{(1,a_1),\ldots,(n,a_n)\}$.
Each instance $a$ in $S$ is associated with a unique 
label/class/prediction $\hat{a}$ belonging to a finite set $C$ (in the running example, $C=\{High, Medium, Low\}$).
$S$ could be simply given via a csv file or via a machine learning process providing the final label/class/prediction.

Next, we rewrite some classical definitions to align with this set-theoretic setting. To illustrate the definition, we use the examples of Table~\ref{example1}, inspired by the COMPAS dataset (Correctional Offender Management Profiling for Alternative Sanctions)~\cite{compas_dataset}, as the running example in this paper. 

COMPAS  is commonly used to analyze bias in criminal justice algorithms. To aid in our explanations, we extracted six features from this dataset. Feature `sex' has two values; `age' has three values ($<25$, 25-45, $>45$);
`race' has six values (African, Asian, Caucasian, Hispanic, Native American, Other). `degree' has two values (M - Misdemeanor, F - Felony); recid (recidivism or likely to re-offend) has two values, and score, i.e., risk of recidive, has three values (Low, Medium and High). Note that the score will be treated as the label or class. Individual `a' with minor offenses $(degree,M)$ is classified as having ``Medium" risk, while individual `c', who has committed a more serious crime $(degree, F)$ and a history of repeated offenses ($recid,Yes$), is classified as ``High" risk.

\begin{definition}\label{projection}
The projection operator $proj$  defined on $Cons(X)$ and leading to $2^{[1,n]}$ is defined as follows:
                      $$\{(f_1,v_1),\ldots, (f_k,v_k)\} \rightarrow \{f_1,\ldots, f_k\}$$
\end{definition}
Roughly speaking, in a consistent set of literals, the projection discards the values while retaining only the corresponding features. We are now in a position to redefine classical concepts according to this setting.
\begin{definition}\label{definition_dag}
Given  $a, b \in Inst(X)$, \\
Agreement set:  $Ag(a,b)=proj(a \cap b) \in [1,n]$ \\
%$Ag(a,b)=\{i \in [1,n]|a_i=b_i \} \in [1,n]$ $Dag(a,b)=\{i \in [1,n]|a_i \neq b_i\} \in [1,n]$
Disagreement set:  $Dag(a,b)= proj(a\setminus b) \in [1,n]$ \\
Hamming distance: $H(a,b)=|Dag(a,b)| = n - |Ag(a,b)|$\\
%Hamming circle: $C(a,r)=\{b \!\in\! Inst(X)| H(a,b)=r\},$ $ r \in [0,n]$\\
$S_a=\{ b \in S, | \hat{b}\neq \hat{a}\}$ (subset of instances in $S$ which do not have label $\hat{a}$)
\end{definition}
Obviously, $Ag(a,b) \cup Dag(a,b) =[1,n]$ and $|Ag(a,b)|+|Dag(a,b)| = n$.\\ 
\\ \noindent{\bf Running example from Table \ref{example1}}:
\begin{itemize}
    \item $a \cap b= \{(age,<25),(degree,M),(recid,No)\}$
    \item $Ag(a,b)=\{age,degree,recid\}$
    \item $Dag(a,b)=\{sex,race\}$
    \item $S_a=\{b,c,g,h\}$, $S_b=\{a,c,d,e,f,h\}$
\end{itemize}
A question that asks ``Why $a$ has label $\hat{a}$" is called {\bf factual question} and will be denoted as $FQ_{S,a}$.
A question that asks ``How can the label of $a$ differ from $\hat{a}$" is called {\bf counterfactual question} and will be denoted as $CFQ_{S,a}$.
\begin{table}[!ht]
    \centering
    \begin{tabular}{|c|c|c|c|c|c||c|}
    \hline 
     & sex  &  age & race & degree & recid & score \\
    \hline
    a &  male       & $<25$  & Caucasian  & M   & No    &  Med \\
    \hline
    b &  female     & $<25$  & African    & M   & No    &  Low \\
    \hline
    c &  male       & $>45$  & African    & F   & Yes   &  High \\
    \hline
    d &  female     & $<25$  & Asian  & F   & No    &  Med \\
    \hline
    e &  female     & $25-45$  & Hispanic  & F  & No    &  Med \\
    \hline
    f &  female     & $<25$  & Caucasian  & F   & No    &  Med \\
    \hline
    g &  female     & $>45$  & Caucasian  & M   & No    &  Low \\
    \hline
    h &  male       & $>45$  & African      & M   & Yes   &  High \\ 
    \hline
    \end{tabular}
    \caption{Running example - inspired by the COMPAS dataset}
    \label{example1}
\end{table}
\subsection{Factual explanations}
A factual explanation for $FQ_{S,a}$ is defined as follows:
\begin{definition} 
A {\bf factual explanation} for $FQ_{S,a}$ is an element $\psi$ of $Cons(X)$ satisfying the following property:
    $$\psi \subset a \mbox{ and } \forall b \in S, \psi \subset b \implies \hat{a}=\hat{b} \quad \quad (1)$$
\end{definition}
It is convenient to denote $FExp_{S,a}$ the set of factual explanations for $FQ_{S,a}$. This definition is quite classical and exactly corresponds to {\it abductive explanations} as formally described in \cite{IgnNarMarAAAI2019,AmgCooECAI2024,JoaoNEURIPS2019,JoaoExplanationsICML2021,GuiMonGiaPedRugTurIEEEIS2019}). Note that:
\begin{itemize}
\item If the condition $\psi \subset a $ is removed, $\psi$ serves as a {\it global} explanation (w.r.t. $S$) for membership in the class $\hat{a} $.
\item If removing any element from $\psi$ causes property (1) to no longer hold, then $\psi$ is {\it minimal}, and it corresponds to a prime implicant (w.r.t. $S$) in logical terms.
\item If we replace $S$ with the whole instance set $Inst(X)$, $\psi$ becomes {\it absolute} in the sense of \cite{JoaoNEURIPS2019}. 
\item If $a $ is disjoint from all elements of $S $, every subset of $a $ constitutes a factual explanation (just because the condition $\psi \subset b)$ is never satisfied in property (1).
\item Additionally, in order to accommodate exceptions, the universal quantification on $b$ over $S$ might be relaxed. Nevertheless, all the works cited above define explanations that do not accommodate exceptions.
\end{itemize}
Obviously, to improve human interpretability, a smaller $\psi $ is generally preferred.\\
\\ \noindent{\bf Running example from Table \ref{example1}}:\\
If every individual $x \in S$ satisfying $\psi=\{(sex,male),(race,Caucasian)\}$ is such that 
$\hat{x}=Med$, then $\psi$ is a factual explanation for $a$. In natural language, if you are a male Caucasian, then your risk of re-offending is medium.
In the following subsection, we investigate counterfactual explanations.
\subsection{Counterfactual explanations}
Roughly speaking, a counterfactual explanation describes some changes in $a$ that will lead to a change in $\hat{a}$, i.e.,
what features to change and how to change them to get a different outcome.
As such, for a given question, we may have different counterfactual explanations, entirely distinct (i.e.,  empty intersection), only comparable in terms of size.
\begin{definition}
A {\bf counterfactual explanation} for $CFQ_{S,a}$ is an element $\psi$ of $Cons(X)$ satisfying the following property:
$$\exists b \in S_a \mbox{ such that }  \psi = b \setminus a \quad (2)$$ 
\end{definition}
Then $\psi$ is the list of literals $(feature,value)$ where $a$ and $b$ differ, ``justifying from observations" why $a$ is not labelled 
$\hat{b}$. In fact, the size of $\psi$, i.e., the number of literals it contains, is exactly $H(a,b)$ and it tells a number of changes to perform on $a$ to change its label. Note that $\psi$ is generally not a  factual explanation for $FQ_{S,b}$. 
Obviously, a counterfactual explanation $\psi$ for $a$ is associated to exactly one unique element $b \in S$ (let us call it a counterfactual example): $\psi$ is telling us what to change in $a$ for $a$ to become exactly $b$. This kind of association is not relevant for factual explanation. It is convenient to denote $CFExp_{S,a}$ the set of counterfactual explanations for $CFQ_{S,a}$.\\
\\ \noindent{\bf Running example from Table \ref{example1}}:\\
For $a$, two counterfactual examples are $b$ with $H(a,b)=2$ and $c$ with $H(a,c)=4$.
The corresponding counterfactual explanations extracted are
\begin{itemize}
    \item $\psi_b=\{(sex,female),(race,African)\}$ from $b$.
    \item $\psi_c=\{(age,>45),(race,African),(degree,F),$ \\$(recid,Yes)\}$ from $c$
\end{itemize}
In the following section, we investigate some properties of counterfactuals which are derived from the definition.
\section{Properties of counterfactual explanations}\label{properties}
We can easily deduce certain properties from the definition.
\begin{property}
    \hfill
    \begin{enumerate}
    \item[i)] If $\psi \in CFExp_{S,a}$ then $\psi \cap a=\emptyset$.
    \item[ii)] If $\psi \in CFExp_{S,a}$ then $a \cup \psi$ is inconsistent.
    \item[iii)] $CFExp_{S,a}=\emptyset$ iff $S_a=\emptyset$. 
    \end{enumerate}
\end{property}
\noindent{\bf Proof:}\\ 
i) Obvious because by definition $\psi = b \setminus a$. This emphasizes that a counterfactual explanation is fundamentally the dual of a factual explanation: for $\psi$ to serve as a factual explanation, it must satisfy $\psi \subset a$.\\
ii) Just because $\psi$ contains literals $l=(j,v_j)$\ such that $l'=(j,v'_j) \in a$ with $v_j \neq v'_j$.\\
iii) $S_a=\emptyset$ means that $S$ is constituted with instances belonging to the same class $\hat{a}$. Then no counterfactual example for $a$ can be found in $S$. Reversely, if $\exists b \in S_a$, then necessarily $b\neq a$ and $\psi=b \setminus a$ is a counterfactual explanation for $CFQ_{S,a}$.\hfill $\Box$\\
\\ \noindent{\bf Running example from Table \ref{example1}}:\\
For $a$, a counterfactual example is $b$ with $H(a,b)=2$.
The corresponding counterfactual explanation extracted is
$\psi_b=\{(sex,female),(race,African)\}$. 
\begin{itemize}
    \item[i)] Obviously $a \cap \psi_b=\emptyset$.
    \item[ii)] $\{(sex,male),(sex,female)\} \subset a \cup \psi_b$ then $a \cup \psi_b$ is inconsistent.
\end{itemize}
The behavior of counterfactual explanations in relation to the sample $S$ is described below.
\begin{property} ({\bf Monotony})
$S \subseteq T \implies CFExp_{S,a}  \subseteq CFExp_{T,a}$.
\end{property}
\noindent{\bf Proof:}\\ 
$S \subseteq T \implies S_a \subseteq T_a$ then $b \in S_a$ implies $b \in T_a$: a counterfactual $\psi$ for $Q_{S,a}$ associated to $b \in S_a$ remains valid since $b \in T_a$ and $\psi$ is still counterfactual for $CFQ_{T,a}$. \hfill $\Box$\\
As previously mentioned, counterfactuals are in a way dual to factual explanations. This can be formalized by the following property.
\begin{property}
({\bf Duality})    Let $a$ be in $S$. \\
    $\forall \psi \in FExp_{S,a}, \forall \psi' \in CFExp_{S,a}, \psi \cup \psi' \mbox{ is inconsistent}$.
\end{property}
\noindent{\bf Proof:}\\ 
Given $\psi'$ counterfactual for $CFQ_{S,a}$, there exists $b \in S$ such that $\hat{b}\neq \hat{a}$  and $\psi'$ is just $b\setminus a$ i.e. $\psi'$ contains all literals from $b$ whose associated value is different in $a$. 
Because $\hat{b}\neq \hat{a}$, whatever $\psi \in FExp_{S,a}$, $\psi \not\subset b$ (remember that $\psi$ is a sufficient condition to have label $\hat{a}$): 
then $\exists l=(j,v_j) \in \psi$ such that $l \notin b$. Let us focus on the value of feature $j$ in $b$: necessarily this value $v'_j$ is such that $v_j \neq v'_j$. Then literal $l'=(j,v'_j) \in \psi'= b \setminus a$. 
Finally, $\psi \cup \psi'$ contains $l$ and $l'$, and is inconsistent.\hfill $\Box$\\
\\ \noindent{\bf Running example from Table \ref{example1}}:\\
We observe that:
\begin{itemize}
    \item $\psi=\{(age,<25),(race,Caucasian)\}$ is a factual explanation for $a$. 
    \item $\psi'_b=\{(sex,female),(race,African)\}$ is a counterfactual for $a$ extracted from $b$.
\end{itemize}
But $\psi \cup \psi'_b= \{(age,<25),(race,Caucasian),$\\
$(sex,female),(race,African)\}$ is obviously inconsistent.

This result can be viewed as a generalization of \cite{JoaoNEURIPS2019}, which focuses solely on the entire set of instances Inst(X) even though certain theoretical instances may not correspond to any real-world item.
It is clear that, for a given instance $a$, multiple counterfactuals may exist. Consequently, it becomes crucial to rank them in order to identify and present ``the best one" to the end user. We investigate the options in the following section.
\section{Ranking counterfactual explanations}\label{ranking}
In fact, any element $b \in S $ with $\hat{b} \neq \hat{a} $ can serve as a basis for counterfactual explanation for
$CFQ_{S,a}$. However, this represents an overly simplistic view of counterfactual explanation. Research indicates that users are motivated to seek counterfactual explanations when aiming for high-quality solutions~\cite{shang2022}. However, if these explanations demand significant cognitive effort, users' trust in the system may diminish~\cite{zhou2017}. To reduce the cognitive load, we propose to display the optimal counterfactual example. To do so, these counterfactuals must be ranked according to a specific criteria, which we explore in the following subsections.
\subsection{Minimal counterfactual explanations}
In fact, we are interested in the minimal number of changes in the feature values of $a$ that would change the label, i.e., counterfactual explanation extracted from a $b$ such that $H(a,b)$ is minimal. 
It is of no help if an instance $b$ very far (w.r.t. Hamming distance) from $a$ is in a distinct class. To provide a relevant counterfactual information about $a$ label, we have to find a $b$ not too distant from $a$ but being in a different class. This is where the notion of minimal counterfactual explanation comes into play (this is also referred as sparsity in \cite{KeaSmyICCBR2020}).
\begin{definition} Let us denote $min_{S,a}=min(\{H(a,b)| b \in S_a\}$.
A minimal counterfactual explanation for $CFQ_{S,a}$ is a counterfactual explanation $\psi_b$ such that $Hamming(a,b) = min_{S,a}$.
\end{definition}
This is equivalent to say that $\psi_b$ is minimal in terms of cardinality. In the terminology of \cite{KeaSmyICCBR2020}, $b$ is a {\it nearest unlike neighbor} of $a$.
Another candidate property for a counterfactual $\psi$ is to be 
{\bf irreducible}.
\begin{definition}({\bf Irreducibility})
    $\psi \in CFExp_{S,a}$ is irreducible iff $$\forall l \in \psi, \exists b \in S \mbox{ s.t. } \hat{a} = \hat{b} \mbox{ and } \psi \setminus {l} \subset b$$
\end{definition}
This tells us that every literal $l=(i,v_i) \in \psi$ is mandatory to ensure the class change.
However, irreducibility says nothing about the Hamming distance between $a$ and $b$: this distance can be high or low.\\
\noindent{\bf Running example from Table \ref{example1}}:\\
We observe that:
\begin{itemize}
    \item $\psi_b$ is a minimal counterfactual for $a$ with $H(a,b)=2$. The other minimal counterfactual for $a$ is $g$.
    \item $\psi_c=\{(age,>45),(race,African),(degree,F),$\\$(recid,Yes)\}$ is a counterfactual extracted from $c$ with $H(a,c)=4$.
    \item However, considering counterexample instance $h$\\$=\{(sex,male),(age,>45),(race,African),$\\$(degree,M),(recid,Yes)\}$ with $score=Low$, then $\psi_c$ is reducible to\\
    $\psi_h=\{(age,>45),(race,African),(recid,Yes)\}$, as  
    $h$ is still a counterfactual example for $a$ with $H(a,h)=3$. 
\end{itemize}
\begin{property}
    If $\psi$ is a minimal counterfactual for $Q_{S,a}$, then $\psi$ is irreducible. The converse, however, is not true.
\end{property}
\noindent{\bf Proof:}\\ 
If $\psi$ counterfactual, associated to counterexample $b$, is minimal, it means there is no counterexample $c$ such that $H(a,c)< H(a,b)=|\psi|$. Removing an element of $\psi$ will lead to a counterexample $c$ such that $H(a,c)=|\psi|-1$ which contradicts the assumption on $\psi$. The converse is not true because we may have irreducible counterexamples which are not minimal: for instance, if the closest counterexamples are at distance $k$, and the other counterexamples are all at distance $k+2$, none of these counterexamples can be reduced to lead to a reduced explanation.
\hfill $\Box$\\
The simple examples given in Table~\ref{example1} are for illustrations only. Real datasets contain more instances, therefore it is likely that there are multiple minimal counterfactuals. We explore this 
possibility in the next section.
\subsection{Optimal counterfactual explanations} \label{optimal_counterfactual}
Still, from experience, multiple minimal counterfactual explanations may exist, but they are not necessarily equivalent. To determine the most appropriate one, it is essential to rank these counterfactuals and identify the `best' option. Below, we outline two potential approaches for achieving this ranking:
\begin{enumerate}
\item {\bf Using Disagreement Sets ($Dag$-based)}:  
   A minimal counterfactual $b$ for an instance $a$ is associated with a disagreement set $Dag(a, b)$. For $k$ minimal counterfactuals, there can be up to $k$ distinct disagreement sets. If a minimal counterfactual has a frequent disagreement set (i.e., one shared by multiple counterfactuals), it can be considered better than a counterfactual with a rare disagreement set. However, in practice, our experiments reveal that for non-binary features, each minimal counterfactual typically has a unique disagreement set (i.e., all disagreement sets are distinct). As a result, this method does not provide a practical way to rank minimal counterfactuals.  
\item {\bf Using Counterfactual Power}:  
%   \textcolor{blue}{A minimal counterfactual $b$ for a given instance $a$ can also serve as a counterfactual for other instances, provided they share the same Hamming distance $H(a, b)$.}
   A minimal counterfactual $b$ for a given instance $a$ can also serve as a counterfactual for other local instances within the same Hamming distance $H(a, b)$.
   In this context, the counterfactual power of $b$ is the number of local instances it can counterfactually explain. Intuitively, this counterfactual power measures the local reusability of that instance. Minimal counterfactuals of $a$ can then be ranked based on how reusable they are in the neighborhood of $a$.
   Our findings indicate that this approach effectively ranks minimal counterfactuals and yields practical insights. 
   To formalize this idea, let us define the hyperball $B(a,b)=\{x \in S| H(x,b)\leq H(a,b)\}$. Obviously 
   $a \in B(a,b)$ but we may have other instances in $B(a,b)$, not in $\hat{b}$, candidate to be counterfactually explained with $b$.
\end{enumerate}
%\textcolor{blue}{\begin{definition}Given $b \in S_a$ a minimal counterfactual example for $a$, let $cf_S(b,a)=S_b \cap C(b,H(a,b))$: the cardinality of the set $cf_S(b,a)$ is the counterfactual power of $b$ w.r.t. $a$.    \end{definition}}
\begin{definition}
Given $b \in S_a$, a minimal counterfactual example for $a$, the counterfactual power of $b$ w.r.t. $a$ is defined as: $$cf_S(b,a)=|S_b \cap B(a,b)|$$    
\end{definition}
In other words, $cf_S(b,a)$ is the number of elements in the hyperball $B(a,b)$ centered at $b$, for which $b$ could be considered as a counterfactual example. Obviously, $cf_S(b,a)\geq 1$ because $a \in S_b \cap B(a,b)$.
This counterfactual power is the notion we are looking for in order to rank the minimal counterfactuals for $a$. 
\\ \noindent{\bf Running example from Table \ref{example1}}:\\
When $S$ is reduced to Table \ref{example1}, we observe that:
\begin{itemize}
    \item $b$ and $g$ are the 2 minimal counterfactuals for $a$ with $H(a,b)=H(a,g)=2$.
    \item $cf_S(b,a)=3$ because $b$ is counterfactual of $\{a,d,f\}$ at distance 2.
    \item $cf(S,g,a)=2$ because $g$ is counterfactual of $\{a,f\}$ at distance 2.
    \item $b$ is the optimal counterfactual.
\end{itemize}
Roughly speaking, increasing the sample set leads to more counterfactual explanations, but obviously a minimal one for $S$ may not be minimal for $T$ as soon as $S \subseteq T$, because $S \subseteq T \implies S_a \subseteq T_a \implies min_{T,a} \leq min_{S,a}$. This applies to optimal counterfactual which can change when we increase the sample.
\subsection{Typicality, capacity and universality metrics}\label{sec:threemetrics}
Beyond human expert evaluation, we can assess the quality of a counterfactual explanation using the three metrics described below. This approach aligns with recent works \cite{Singh2021} that emphasize the importance of evaluating counterfactual explanations using multiple metrics to ensure their effectiveness and reliability.
Let $a \in S$, $b$ a minimal counterfactual for $a$. 
Let us consider the hyperball $B(a,b)$ centered at $b$ with radius $H(a,b)$ i.e.  $B(a,b)=\{x \in S| H(x,b)\leq H(a,b)\}$.
\begin{itemize}
    \item This hyperball may contain instances belonging to the same class as $b$, which are also counterfactuals for $a$, though not necessarily minimal. If this hyperball includes a significant proportion of instances labeled $\hat{b}$, it suggests that $b$ is relatively typical of its class. We define {\bf typicality} as the ratio of such instances within the hyperball to the number of instances labeled $\hat{b}$:
    $$\frac{|\{x \in S| \hat{x}=\hat{b} \} \cap B(a,b)|}{|\{x \in S| \hat{x}=\hat{b}\}|}$$
 \item We also assess the capacity of $b$ to counter-explain elements within the hyperball. Given that $b$ is a candidate to counter-explain all elements in the hyperball not in $\hat{b}$, the proportion of such elements can serve as a good indicator of this capacity. We define {\bf capacity} as the ratio of instances not labeled $\hat{b}$ to the total number of instances within the hyperball:
$$\frac{|\{x \in S| \hat{x} \neq \hat{b} \} \cap B(a,b)|}{|B(a,b)|}=\frac{|S_b \cap B(a,b)|}{|B(a,b)|}=\frac{|cf_S(a,b)|}{|B(a,b)|}$$
\item Now, if the hyperball contains a significant number of instances labeled $\hat{a}$, then $b$ serves as a counterfactual for these instances, although it may not be the optimal one. This property imparts a sense of ``universality" to $b$, reflecting its broader applicability across multiple instances with label $\hat{a}$. 
We define {\bf universality} as the proportion of such instances within the hyperball relative to the total number of instances in the ball:
$$\frac{|\{x \in S| \hat{x}=\hat{a} \} \cap B(a,b)|}{|B(a,b)|}$$
\end{itemize}
When reduced to a case where we have only two classes, capacity and universality are identical. 
The values of typicality, capacity, and universality follow the principle: the higher, the better! 
But obviously, these values for the optimal counterfactual may lack absolute significance and must be evaluated in comparison to the typicality, capacity and universality of a randomly selected counterfactual from the remaining minimal counterfactuals.
By comparing these values, we aim to provide a more robust assessment of the counterfactuals' quality.\\
%\textcolor{red}{I AM NOT SURE OF THIS -- Note that counterfactual power $cf_S(b,a)$ could also be defined as a ratio like 
%$$\frac{|\{x \in S| \hat{x} \neq \hat{b} \} \cap B(a,b)|}{|B(a,b)|}=\frac{|S_b \cap B(a,b)|}{|B(a,b)|}=capacity(b)$$
% but this will not change the ranking of the minimal counterfactuals and the optimal one will still be the same.\\
% Instead of counting a number of values (absolute value) we could have evaluated a ratio (relative value) ???
%}
\section{Experiments}\label{expes}
We have conducted the experiments using 12 categorical datasets shown in Table \ref{real_datasets}. These data are
from OpenML (https://www.openml.org/), mlData.io (https://www.apispreadsheets.com/datasets) and mlr3fairness (https://mlr3fairness.mlr-org.com) which are open platforms for sharing datasets, algorithms, and experiments.  All of the datasets are categorical, with non binary features, with more than 6 dimensions and more than 1000 rows.
\begin{table}[!ht]
    \centering
    \begin{tabular}{|c|c|c|c|}
    \hline
         Name & dimension & classes & rows \\
         \hline
         Adult & 14 & 2 & 44842 \\
         \hline
         Bach & 14 & 102 &5665 \\
         \hline
         Cars & 6 & 4 &1728 \\
         \hline
         Chess & 6 & 18 &28056 \\
         \hline
         Contraception & 9 & 3 & 1473 \\
         \hline
         Mushrooms & 22 & 2 & 8416 \\
         \hline
         Phishing & 30 & 2 & 11055 \\
         \hline
         Portugal bank & 16 & 2 & 4521 \\
         \hline     
         Compas & 14 & 3 & 4513 \\
         \hline
         Loan & 14 & 3 & 8848 \\
         \hline
         Marketing & 15 & 2 & 7842 \\
         \hline
         Retention & 32 &  3 &  4424 \\
         \hline
    \end{tabular}
    \caption{The 12 real datasets used in the experiments}
    \label{real_datasets}
\end{table} 
\subsection{Counterfactuals versus Hamming distance}
\begin{figure}[!ht]
    \centering
       \includegraphics[width=0.3\columnwidth]{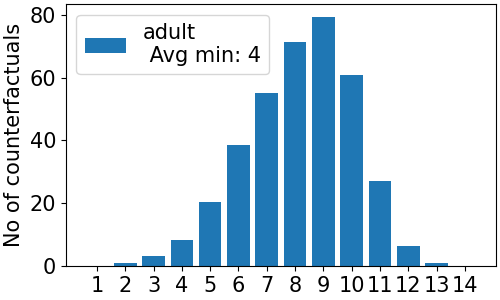}
        \label{fig:sub1}
    \hfill
        \includegraphics[width=0.3\columnwidth]{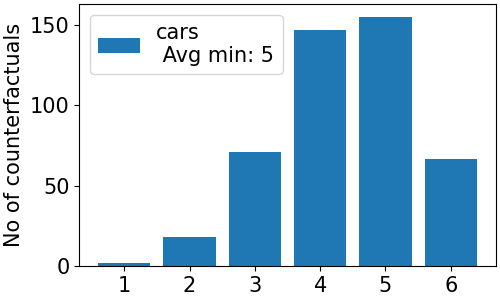}
        \label{fig:sub2}
    \hfill
        \includegraphics[width=0.3\columnwidth]{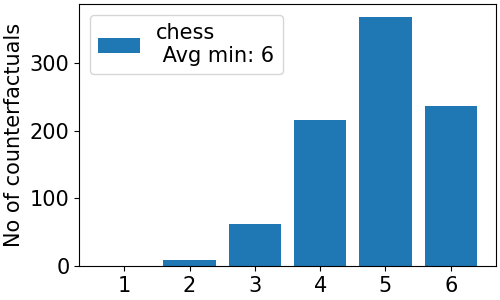}
        \label{fig:sub3}

    \centering
       \includegraphics[width=0.3\columnwidth]{images/chess.png}
        \label{fig:sub4}
    \hfill
        \includegraphics[width=0.3\columnwidth]{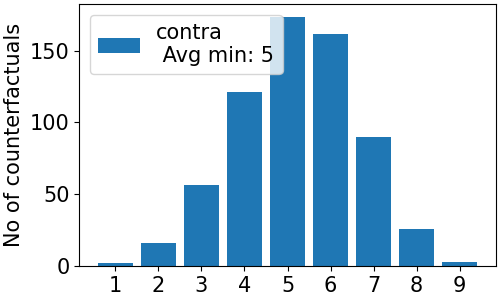}
        \label{fig:sub5}
    \hfill
        \includegraphics[width=0.3\columnwidth]{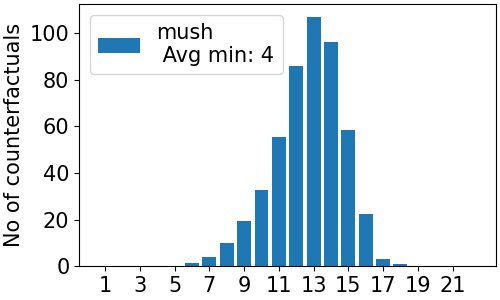}
        \label{fig:sub6}

    \includegraphics[width=0.3\columnwidth]{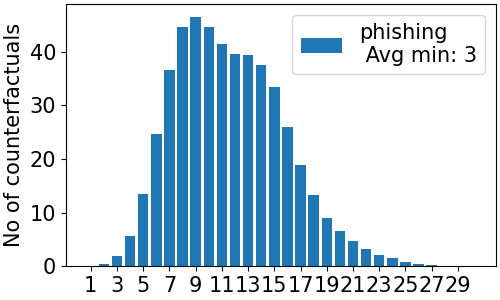}
        \label{fig:sub7}
    \hfill
        \includegraphics[width=0.3\columnwidth]{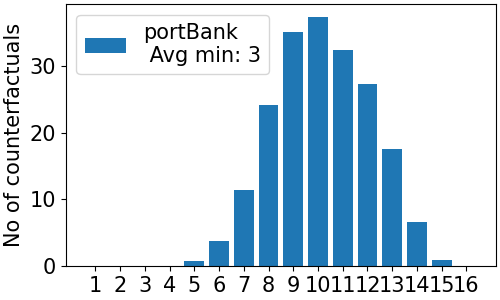}
        \label{fig:sub8}
    \hfill
        \includegraphics[width=0.3\columnwidth]{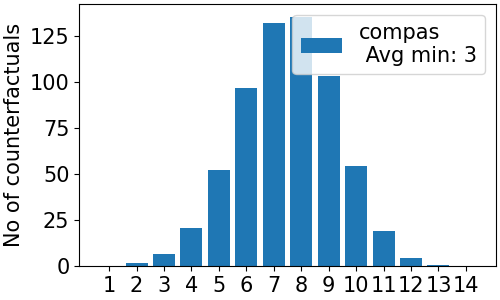}
        \label{fig:sub9}

    \includegraphics[width=0.3\columnwidth]{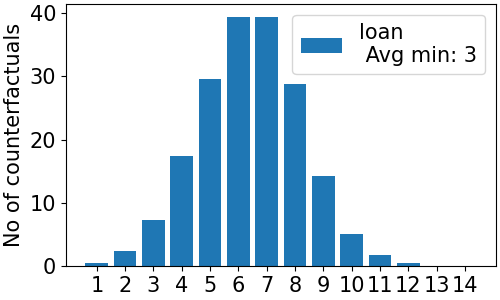}
        \label{fig:sub10}
    \hfill
        \includegraphics[width=0.3\columnwidth]{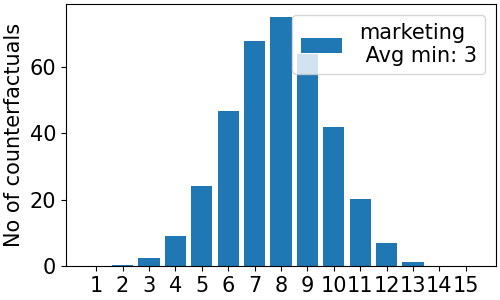}
        \label{fig:sub11}
    \hfill
        \includegraphics[width=0.3\columnwidth]{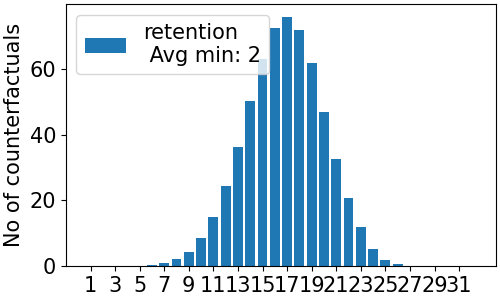}
        \label{fig:sub12}
    \caption{The number of counterfactual for Hamming distance 1 to $n$, where $n$ is the number of features}
    \label{rankingHamming}
\end{figure}

To gain practical insight into the relationship between the Hamming distance and counterfactual explanations, we study the
distribution of counterfactuals corresponding to Hamming distance ranging from $1$ to $n$, where $n$ is the dimension of the dataset. 
To do that, we randomly sample 1,000 instances from each dataset, then calculate the number of counterfactuals per Hamming distance for each instance. Then we average the results over 1,000 and display the findings in Figure \ref{rankingHamming}.
\noindent % Prevent indentation

We observe that the number of counterfactuals increases with Hamming distance but begins to decline once this distance exceeds a certain threshold. In other words, only a few counterfactuals are either very close or very far from the given instance. This figure illustrates that, on average, each instance in all datasets has at least two minimal counterfactuals going up to 6 for the chess dataset! Note that for counterfactual explanation, we would like to display only one counter example to the end user. As explained in Section~\ref{optimal_counterfactual}, we achieve this by ranking the counterfactuals, and we show the effectiveness of this approach in the next section.

\subsection{Minimal counterfactuals ranking}\label{sec:minimal_counterfactuals_ranking}
\begin{figure}[!ht]
    \centering
    \includegraphics[width=0.8\linewidth]{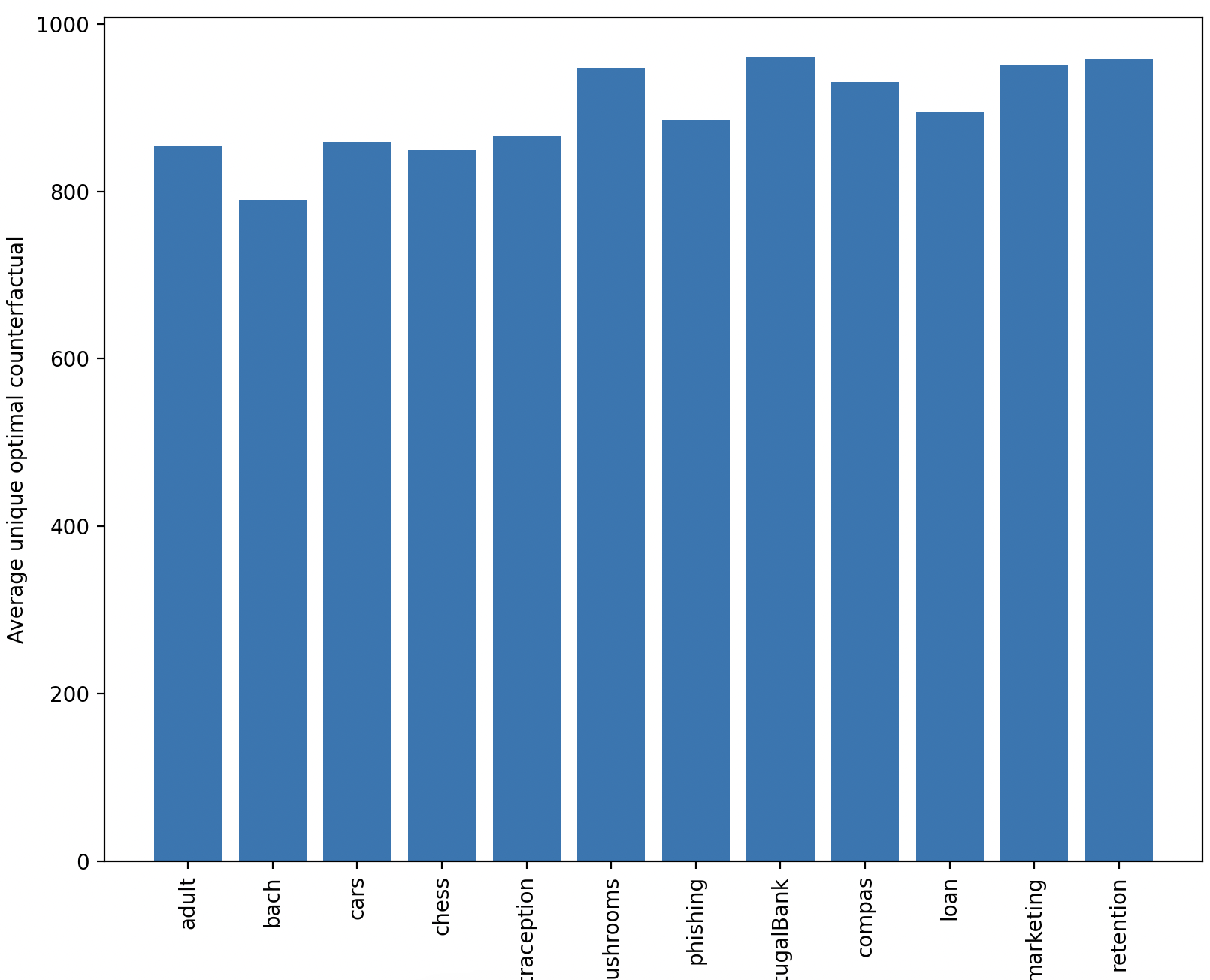}
    \caption{Average proportion of unique optimal per dataset}
    \label{barChart}
\end{figure}

To highlight the effectiveness of the proposed counterfactual ranking, we estimated how often we can rank the minimal counterfactuals using the following protocol:
\begin{enumerate}
    \item For each dataset, randomly choose sample $S$ of size 1,000. 
    \item For each element of $S$, compute whether there is a unique optimal counterfactual. 
    \item Count how often it occurs among the 1000 elements.
    \item Repeat steps 1 to 4 for 100 times.
    \item Average the results over 100 times.
\end{enumerate}
The results are shown in Figure \ref{barChart}, highlighting that a large majority (usually more than 80\%) of elements have a unique optimal counterfactual in the sample.
\noindent 
To better understand the proposed ranking method, we examine the counterfactual power of each minimal counterfactual. We observed that the counterfactual power of the highest-ranked counterfactual can be very close to that of the second-highest, resulting in a small gap between their values. This small gap may diminish the significance of the ranking, as the highest-ranked counterfactual is not substantially better than the second-highest.

To quantify this, we calculate the average gap in counterfactual power between the highest- and second-highest-ranked counterfactuals, relative to the power of the highest-ranked one. For instance, if the highest-ranked counterfactual explains 45 instances and the second-highest explains 40, the relative gap of 5 is computed as $5/45=1/9 \sim 0.11$. This gap is less significant than a gap of $5$ when the highest-ranked counterfactual has a power of only $10$, resulting in a relative gap of $5/10=0.5$. The closer this ratio is to $1$, the more significant is the highest-ranked counterfactual.

\begin{figure}[!ht]
    \centering
    \includegraphics[width=0.8\linewidth]{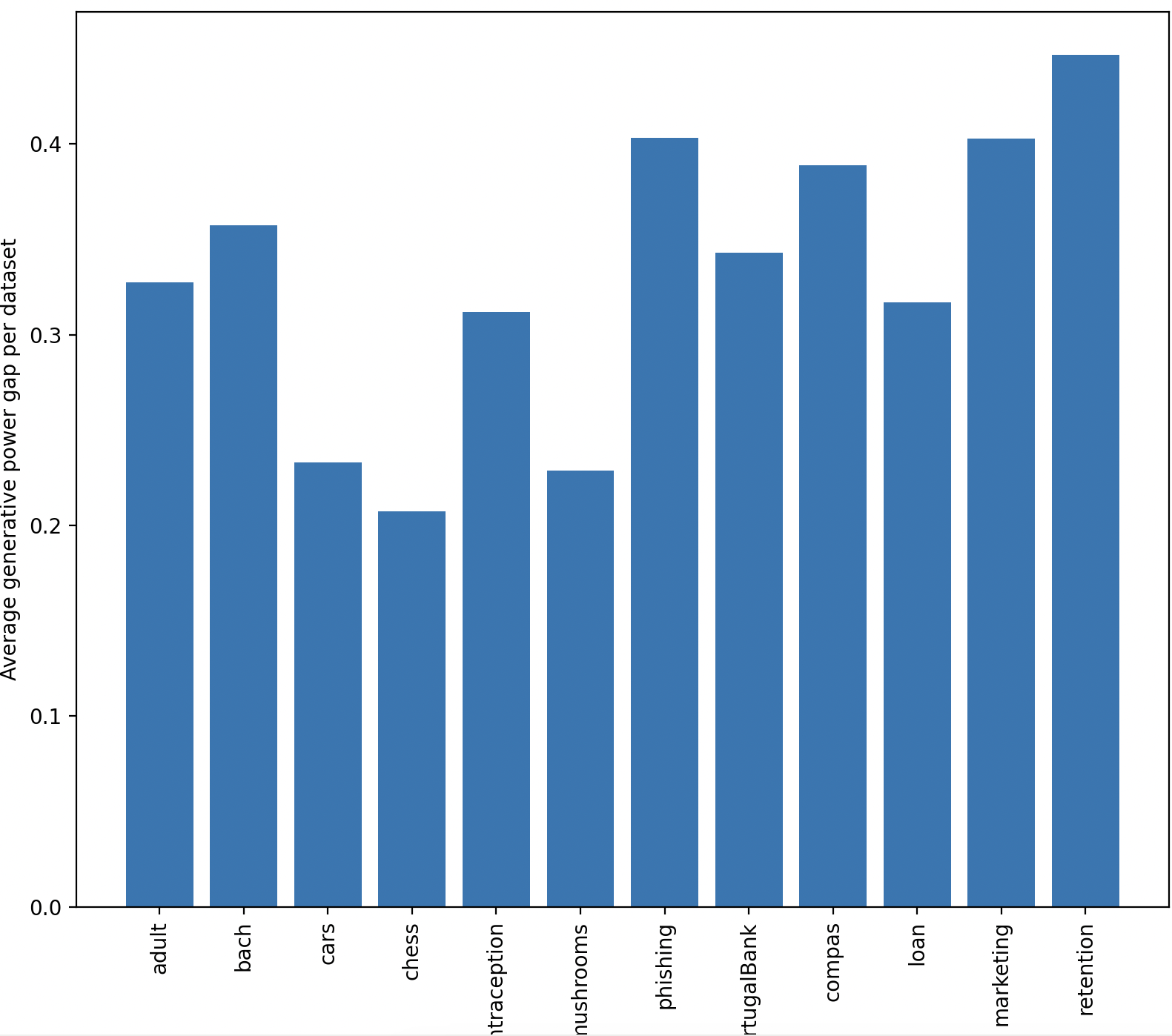}
    \caption{Average gap in counterfactual power}
    \label{average-gap}
\end{figure}

In this experiment, we started from the protocol described in Section~\ref{sec:minimal_counterfactuals_ranking} and we modified steps 2 and 3 as below:
\begin{enumerate}
  \setcounter{enumi}{1}
  \item For each element of $S$, compute whether there is a unique optimal counterfactual and compute the gap between highest and second ranked.
    \item Count how often it happens among the 1000 elements and compute average gap on 1000 elements.  
\end{enumerate}
Figure \ref{average-gap} displays the average value of this gap for the 12 datasets.
We observe that, on average, the relative gap is at least $0.2$, indicating that the counterfactual power of the highest-ranked counterfactual is at least 20\% greater than that of the second-highest-ranked one. This demonstrates that the optimal counterfactual is significantly more effective or ``powerful" than other candidate minimal counterfactuals for a given instance. It also serves as validation for our proposed ranking method.

\subsection{Optimal counterfactual evaluation}
\begin{figure}[!ht]
    \centering
    \includegraphics[width=0.8\linewidth]{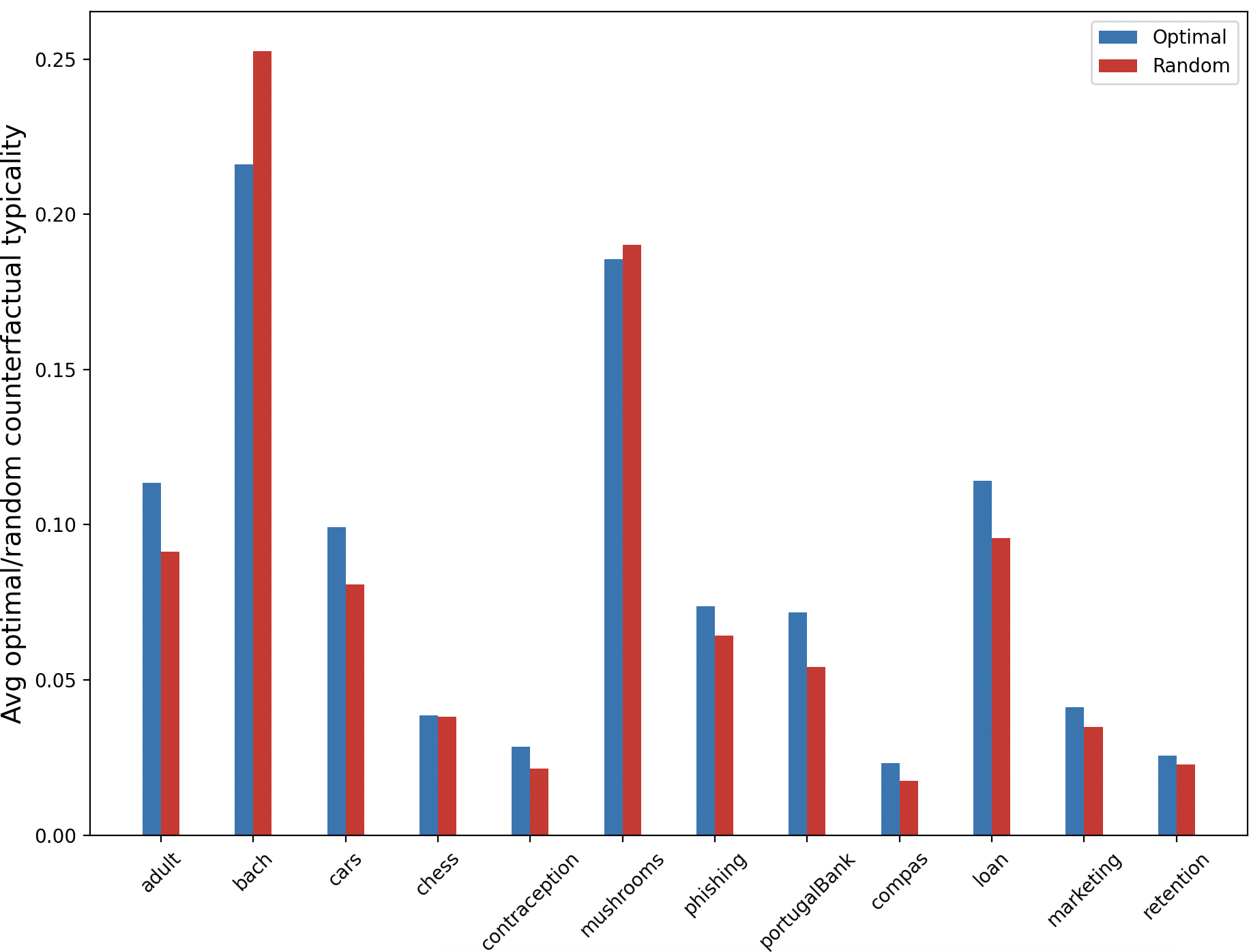}
    \caption{Average typicality: optimal versus random}
    \label{average-typicality}
\end{figure}

\begin{figure}[!ht]
    \centering
    \includegraphics[width=0.6\linewidth]{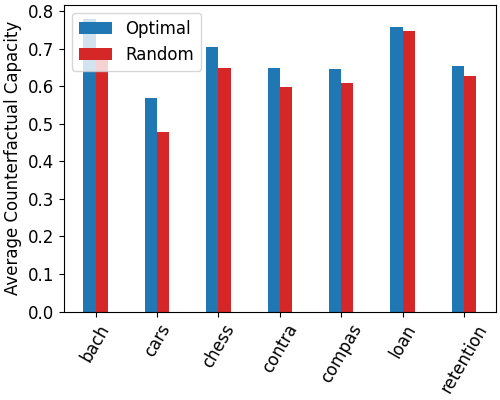}
    \caption{Average capacity: optimal versus random (non binary classes)}
    \label{average-capacity}
\end{figure}

\begin{figure}[!ht]
    \centering
    \includegraphics[width=0.8\linewidth]{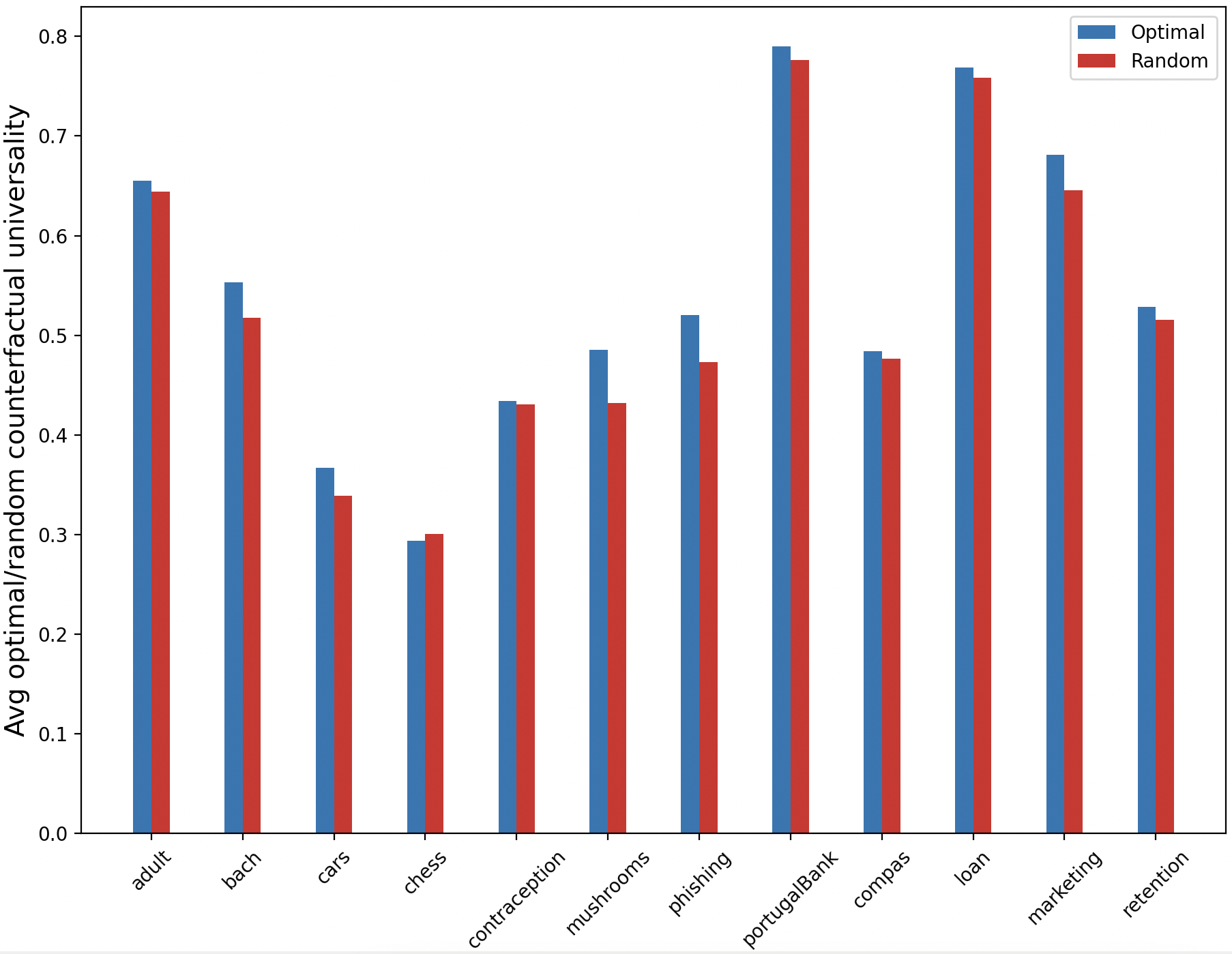}
    \caption{Average universality: optimal versus random}
    \label{average-universality}
\end{figure}

Following the protocol described in Section \ref{sec:minimal_counterfactuals_ranking}, we computed the  typicality, capacity and universality metrics (see Section~\ref{sec:threemetrics}) for each instance in the sample. Metrics were calculated for both the optimal counterfactual and a randomly selected minimal counterfactual. By averaging these values over 100 iterations, we obtained a comprehensive assessment of the ``quality" of our optimal counterfactuals, with the results presented in Figures~\ref{average-typicality}, ~\ref{average-capacity} and~\ref{average-universality}. The blue bars represent the metrics values for the optimal counterfactuals, while the red bars represent the metrics values for a randomly selected minimal counterfactuals. The results clearly show that, most of the time, the optimal counterfactuals are more ``typical", ``capable" (when more than 2 classes) and ``universal" than their randomly chosen counterparts.
\subsection{Link between optimal counterfactual and relevant features}
In many cases, it is important to identify the set $\mathcal{R}$ of features relevant to classification or decision-making, allowing us to discard explanations that rely on irrelevant features. Numerous feature selection techniques are readily available off the shelf. However, we may question whether the optimal counterfactual $b$, when it exists, relies exclusively on relevant features to explain the label change from $\hat{a}$.
Two facts seem obvious:
\begin{itemize}
    \item At least one relevant feature must belong to $Dag(a, b)$, as these features are directly associated with the class change and, by definition, $\hat{a} \neq \hat{b}$.
    \item $Dag(a, b)$ likely includes irrelevant features because, when $S$ is smaller than the entire instance space $Inst(X)$, an optimal counterfactual with respect to $S$ may differ from $a$ on more features than strictly necessary. However, if $S = Inst(X)$, then every minimal counterfactual differs only on the set of relevant features.
\end{itemize}
The question could be set as follows: "What is the proportion of relevant features utilized in a counterfactual explanation derived from $b$ represented by the ratio $\frac{|Dag(a,b) \cap \mathcal{R}|}{|\mathcal{R}|}$ ?". 
Clearly, this ratio must be greater than $\frac{1}{|\mathcal{R}|}$ since for two elements $a,b$ of $S$ to belong to distinct classes, it is necessary that $Dag(a,b) \cap \mathcal{R} \neq \emptyset$.

To bring a partial answer to this question, we conducted experiments on  synthetic datasets where the relevant features are explicitly known and they are categorical with 3 candidate values.
The experiments were performed on  four synthetic categorical datasets with a dimensionality of $20$, where the final label is determined by four simple functions involving either 2, 3, 4 or 6 attributes among 20, rendering all other features entirely irrelevant. 
These datasets were sampled with $500, 1000, 2000, 5000, 10000, 20000$ elements, and we have calculated the ratio $\frac{|Dag(a,b) \cap \mathcal{R}|}{|\mathcal{R}|}$, averaging the results to estimate the proportion of relevant features employed in counterfactual explanations.
In all cases, we observe an average ratio of less than $0.60$, indicating that, in general, the optimal counterfactual does not predominantly focus on relevant features. 
Nevertheless, it is worth noting that 20,000 remains a relatively small number compared to the total number of instances in $Inst(X)$ which in this case is $3^{20}$.
Additional experiments are then required to gain deeper insights into the relationship between optimal counterfactuals and relevant features.

\section{Conclusion}\label{conc}
We have proposed a literal-based definition of counterfactual explanations and analyzed its properties, including its connection to factual explanations. 
Observing that multiple counterfactuals may exist for a single instance, we developed a systematic method to rank them, introducing the concept of optimal counterfactuals. Our experiments revealed that, in most cases, a single optimal counterfactual exists. 

Furthermore, we have introduced three metrics (typicality, capacity and universality) to evaluate the quality of the optimal counterfactual.
We found that, in general, for a given instance, not only does our method provide a unique optimal counterfactual explanation, but this optimal counterfactual also outperforms a randomly selected minimal counterfactual with respect to the three evaluation metrics.

Counterfactual explanations can serve as valuable guidance for altering the outcome of a given case. However, certain features, such as sex, race, and age, are immutable and cannot be changed. In such cases, while the counterfactual explanation may help clarify the reasoning behind a decision, it cannot be practically implemented to achieve the desired outcome. As presented here, the concept of optimal counterfactuals is flexible and can be easily adapted to scenarios where certain features are irrelevant or protected, as required under frameworks like the EU GDPR \cite{GDPR2017}.

From a theoretical perspective, a deeper investigation of the three metrics is necessary in terms of probability measures.
Additionally, translating counterfactual explanations into a format that is more intuitive and easily understood by end users will be crucial. This will enable the validation of our approach through human feedback, allowing us to assess whether the provided counterfactual explanations align with their expectations. This effort will be a central focus of our future work.
\newpage
\bibliographystyle{unsrtnat}
\bibliography{biblio}
\end{document}